\documentclass[journal]{IEEEtran}

\usepackage{lineno,hyperref}\usepackage{amsmath}
\usepackage{graphicx}
\usepackage{threeparttable}
\usepackage{multirow}
\usepackage{colortbl}
\usepackage{amssymb}
\usepackage{color, soul}
\usepackage{xcolor}
\begin{document}
\sethlcolor{yellow}
\title{Gland Instance Segmentation Using \\Deep Multichannel Neural Networks}

\author{Yan Xu, Yang Li, Yipei Wang, Mingyuan Liu, Yubo Fan, Maode Lai, and Eric I-Chao Chang*
\thanks{Manuscript submitted for review on December 24, 2016; accepted on March 11, 2017. This work is supported by Microsoft Research under the eHealth program, the Beijing National Science Foundation in China under Grant 4152033, the Technology and Innovation Commission of Shenzhen in China under Grant shenfagai2016-627, Beijing Young Talent Project in China, the Fundamental Research Funds for the Central Universities of China under Grant SKLSDE-2015ZX-27 from the State Key Laboratory of Software Development Environment in Beihang University in China.} 
\thanks{Yan Xu, Yang Li, Yipei Wang, Mingyuan Liu and Yubo Fan are with the State Key Laboratory of Software Development Environment and Key Laboratory of Biomechanics and Mechanobiology of Ministry of Education and Research Institute of Beihang University in Shenzhen, Beihang University, Beijing 100191, China.}
\thanks{Yan Xu and *Eric I-Chao Chang are with the Microsoft Research, Beijing 100080, China (*Corresponding author: echang@microsoft.com).}
\thanks{Maode Lai is with the Department of Pathology, School of Medicine, Zhejiang University, China.}
\thanks{Copyright (c) 2016 IEEE. Personal use of this material is permitted. However, permission to use this material for any other purposes must be obtained from the IEEE by sending an email to pubs-permissions@ieee.org.}

}

\maketitle

\begin{abstract}
Objective: A new image instance segmentation method  is proposed to segment individual glands (instances) in colon histology images. This process is challenging since the glands not only need to be segmented from a complex background, they must also be individually identified.
Methods: We leverage the idea of image-to-image prediction in recent deep learning by designing an algorithm that automatically exploits and fuses complex multichannel information - regional, location and boundary cues - in gland histology images.
Our proposed algorithm, a deep multichannel framework, alleviates heavy feature design due to the use of convolutional neural networks and is able to meet multifarious requirements by altering channels.
Results: Compared to methods reported in the 2015 MICCAI Gland Segmentation Challenge and other currently prevalent instance segmentation methods, we observe state-of-the-art results based on the evaluation metrics.
Conclusion: The proposed deep multichannel algorithm is an effective method for gland instance segmentation.
Significance: The generalization ability of our model not only enable the algorithm to solve gland instance segmentation problems, but the channel is also alternative that can be replaced for a specific task. 
\end{abstract}

\IEEEpeerreviewmaketitle

\section{Introduction}

\IEEEPARstart{E}{xisting} in most organ systems as important structures, glands secrete proteins and carbohydrates. However, adenocarcinomas, the most prevalent type of cancer, arises from the glandular epithelium \cite{travis2011international}. The morphology of glands determines whether they are benign or malignant and the level of severity \cite{nguyen2012structure}. Segmenting glands from the background tissue is important for analyzing and diagnosing histological images.

\begin{figure}[!t]\centering
\includegraphics[width=3.4in]{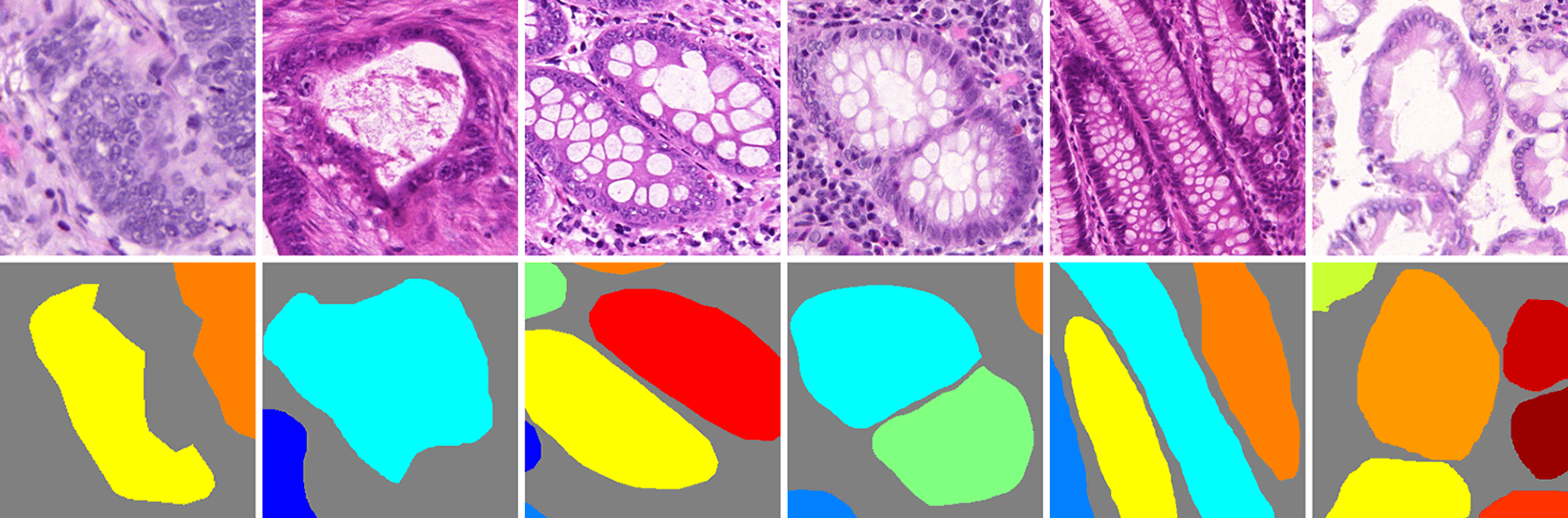}
\caption{Gland Haematoxylin and Eosin (H\&E) stained slides and ground truth labels. Images in the first row exemplify different glandular structures. Characteristics such as heterogeneousness and anisochromasia can be observed in this figure. The second row shows the ground truth. To achieve better visual effects, each color represents an individual glandular structure.}
\label{fig1}
\end{figure}

In gland labeling/segmentation, each pixel is assigned one label to represent whether the pixel belongs to the foreground (gland) or the background. However, which gland the foreground pixel belongs to is still not determined. In order to analyze the morphology of glands, they need to be recognized individually. Each pixel needs to be classified and it must be determined which gland the pixel belongs to, which is to assign a gland ID to each foreground pixel. We call this task as \textbf{gland instance segmentation} (as shown in Fig.~\ref{fig1}). In this paper, we aim to solve the gland instance segmentation problem. We formulate this problem as two subproblems - gland labeling/segmentation \cite{al2010improved,veta2013automatic} and instance recognition.

The intrinsic properties of gland histopathological image pose plenty of challenges in instance segmentation \cite{dimopoulos2014accurate}. First of all, heterogeneous shapes make it difficult to use mathematical shape models to achieve segmentation. As Fig.1 shows, the cytoplasm being filled with mucinogen granule causes the nucleus to be extruded into a flat shape whereas the nucleus appears as a round or oval body after secreting. Second, variability of intra- and extra- cellular matrices often leads to anisochromasia. Therefore, the background portion of histopathological images contains more noise like intensity gradients, compared to natural images.  Several problems arise in our exploration of analyzing gland images: 1) some objects are very close together making only the tiny gaps between them visible when zooming in on a particular image area; or 2) one entity borders another making their edges adhesive to each other. We call this an problem of \emph{`coalescence'}. If these problems are omitted during instance recognition process, even if there is only one pixel coalescing with another, the algorithm will consider two instances as one. 

Gland labeling/segmentation, as one subproblem of gland instance segmentation, is a well-studied field where various methods have been explored, such as morphology-based methods \cite{naik2007gland,nguyen2010automated,naik2008automated,paul2016gland} and graph-based methods \cite{egger2013pcg,tosun2011graph}. However, glands must be recognized individually to enable the following morphology analysis. Gland segmentation is insufficient due to its inability to recognize each gland in histopathological images. MICCAI 2015 Gland Segmentation Challenge Contest \cite{sirinukunwattana2016gland} has drawn attention to gland instance segmentation. The precise gland instance segmentation in histopathological images is essential for morphology assessment, which is proven to be not only a valuable tool for clinical diagnosis but also a prerequisite for cancer grading \cite{fleming2012colorectal}.

Although gland instance segmentation is a relatively new subject, instance segmentation in nature images has attracted much interest from researchers.  
Ever since SDS \cite{hariharan2014simultaneous} raised this problem and proposed a basic framework to solve it, other methods have been proposed thereafter, such as hypercolumn \cite{hariharan2015hypercolumns} and MNC \cite{dai2015instance}, which merely optimize and accelerate the feature extraction process. All of these algorithms fall into a routine that detects objects first and then segments object instances inside the detected bounding boxes.

In medical image analysis, traditional methods are more prevalent for segmenting gland instances instead of learning-based methods. Traditional methods depend heavily on hand-craft features and prior knowledge. In natural images, instance segmentation algorithms are mostly the pipeline of object detection and masking \cite{hariharan2014simultaneous,hariharan2015hypercolumns,dai2015instance}. 
The objects in natural images are regular-shaped, and relatively easy to segment by first creating bounding boxes for each one.  
However, most glands are irregular in shape, which increases the difficulty of detecting the whole gland structure. 
Thus the traditional instance segmentation methods for natural images are not suitable for gland instance segmentation.

In a broad sense, gland instance segmentation can be viewed as gland labeling process with commutative labels. Thus gland labeling can offer useful cues for gland instance segmentation. 
The latest advantages in deep learning technologies have led to explosive growth in machine learning and computer vision for building systems that have shown significant improvements in a huge range of applications such as image classification \cite{krizhevsky2012imagenet,vggnet} and object detection  \cite{girshick2015fast}.
The fully convolutional neural networks (FCN) \cite{long2015fully} permit end-to-end training and testing for image labeling;
holistically-nested edge detector (HED) \cite{xie15hed} detector learns hierarchically embedded multiscale edge fields to account for the low-, mid-, and high- level information for contours and object boundaries; Faster R-CNN \cite{ren2015faster} predicts object locations and compensates for the possible failure of edge prediction.
We solve the gland instance segmentation problem by multitask learning. One task is to segment the gland images, and another task is to identify the gland instances. 
In the gland segmentation subtask, a fully convolutional neural network (FCN) \cite{long2015fully} model is employed to exploit the advantage of end-to-end training and image-to-image prediction. In the gland instance recognition subtask, a holistically-nested edge detector (HED) and a Faster R-CNN object detector are applied to define the instance boundaries. 

\begin{figure}[!t]
\centering
\includegraphics[width=3in]{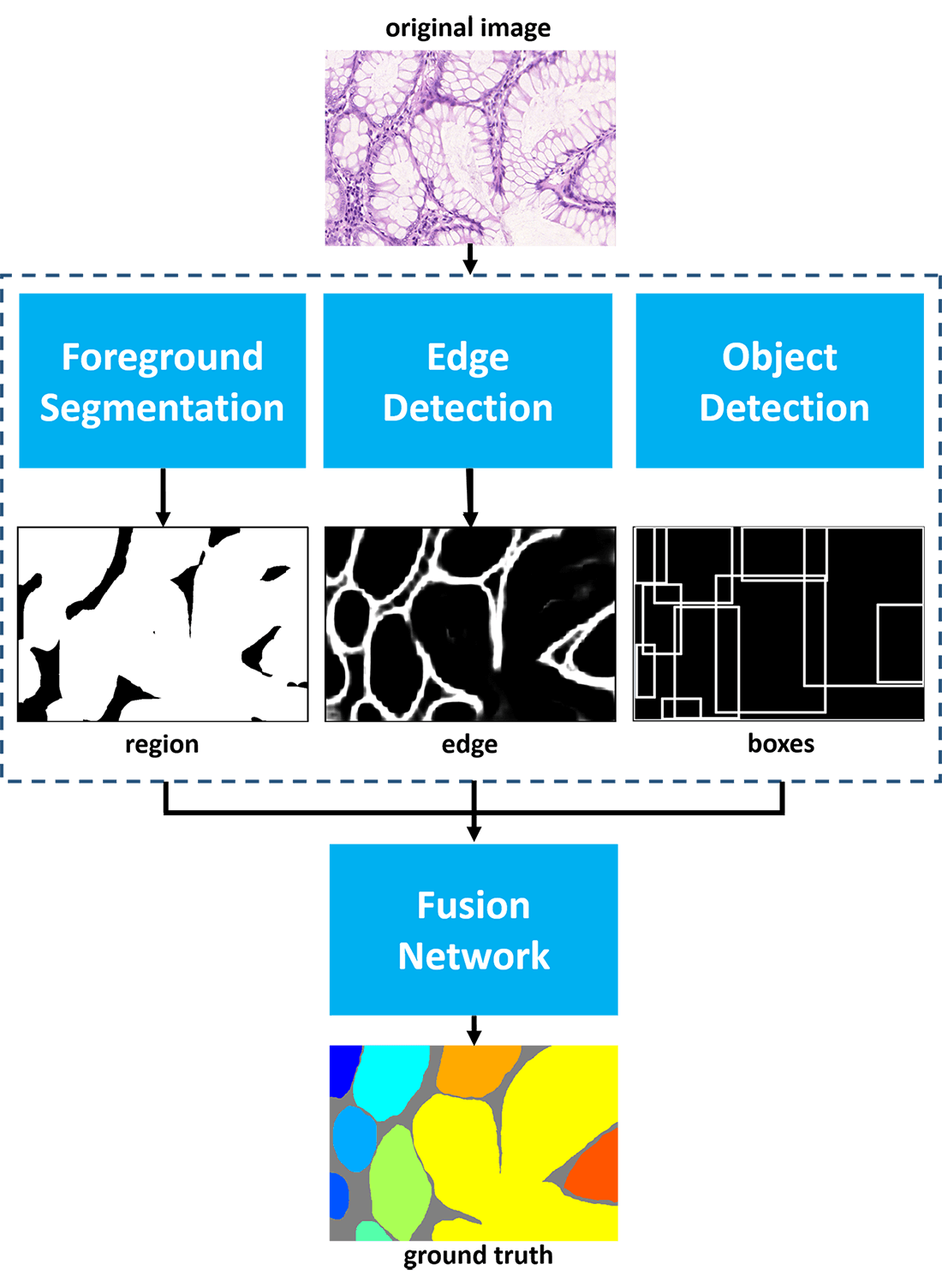}
\caption{This illustrates a brief structure of the proposed algorithm. The foreground segmentation channel distinguishes gland pixels from the background. The edge detection channel outputs the result of boundary detection. The object detection channel detects glands and their regions in the images. A convolution neural network concatenates features generated by different channels and produces segmented instances. The white areas in subimage ``region, edge and boxes'' represent the results of the recognized glands, edges and detected bounding boxes. }
\label{model1}
\end{figure}
We make use of multichannel learning to extract region, boundary and location cues and solve the instance segmentation problem in gland histology images (as shown in Fig.~\ref{model1}). Our algorithm is evaluated on the dataset provided by the MICCAI 2015 Gland Segmentation Challenge Contest \cite{sirinukunwattana2016gland} and achieves state-of-the-art performance among all participants and other popular methods of instance segmentation. We conduct a series of ablation experiments and prove the superiority of the proposed algorithm.

This paper is arranged as follows. We formulate the instance segmentation problem in Section \ref{pro}. Section \ref{related} is a review of related previous works. In section \ref{method}, we describe the complete methodology of the proposed algorithm of gland instance segmentation. Section \ref{exp} is a detailed evaluation of our method. Section \ref{con} summarizes our conclusion.

\section{Problem}
\label{pro}
We formulate the instance segmentation problem by two subproblems, labeling/segmentation and instance recognition.

\begin{figure*}[!t]
\centering
\includegraphics[width=0.8\textwidth]{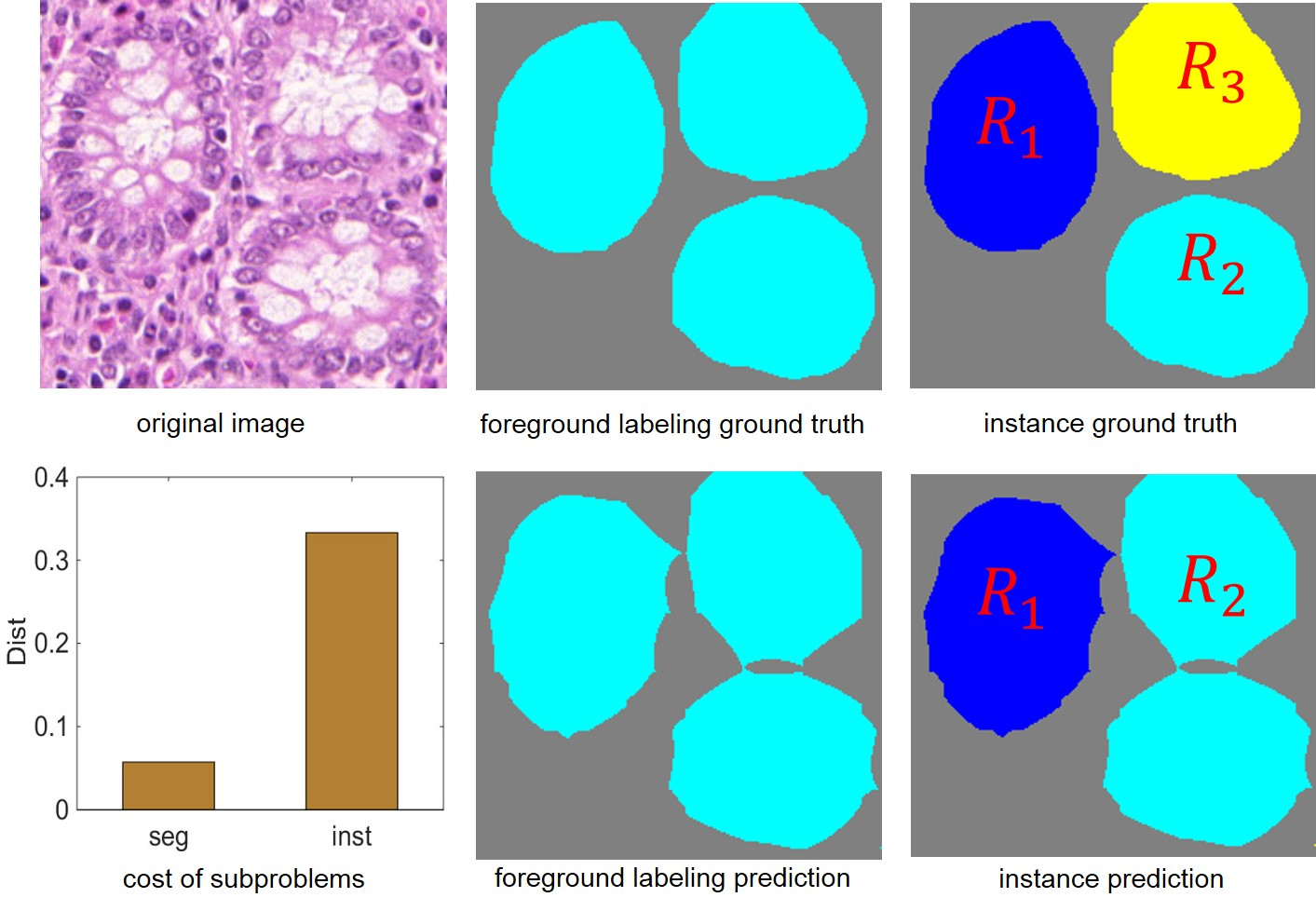}
\caption{This illustrates two subproblems of gland instance segmentation. Gland instance segmentation can be formulated into foreground labeling/segmentation and gland instance recognition two subproblems. As demonstrated in images of the second column, a small amount of prediction errors have little influence on the final cost function for the foreground labeling/segmentation subproblem; however, for the gland instance recognition subproblem, even a few pixels predicted incorrectly can highly increase the cost, which is shown in images of the third column. The bar chart shows the cost of two subproblems.}
	\label{sub}
\end{figure*}

We denote $D=\{(X_{n},Y_{n},Z_{n}),n=1,2,...,N\}$ as the input training dataset, where $N$ is the image amount. We subsequently drop the subscript $n$ for notational simplicity, since we consider each image independently. $X=\{x_{j},j=1,2,...,|X|\}$ denotes the raw input image, $Y=\{y_{j},j=1,2,...,|X|\},{y_{j}}\in\{0,1\}$ denotes the corresponding segmentation label and $Z=\{R_{k},k=0,1,2,...,K\}$ denotes the instance label, in which $R_{k}=\{{(p,q)}\}$ denotes the coordinates set of pixels inside of region $R_{k}$. When $k$ equals 0, it denotes the background area and it denotes the corresponding instance when k takes other values. $K$ is the total instance number. Regions in the image satisfy the following relations:
\begin{equation}
R_{k} \cap R_{t}=\varnothing,\forall k\neq t,
\end{equation}
\begin{equation}
\cup R_{k}=\Omega.
\end{equation}
$\Omega$ denotes the whole image region.
Note that instance labels only count gland instances thus they are commutative. Our objective is to segment glands while ensuring that all instances are differentiated.
Note that the labeling/segmentation subproblem is a binary classification problem. $\hat{Y}$ represents the labeling/segmentation result, the cost function is:
\begin{equation}
Dist(Y,\hat{Y})=\frac{1}{|Y|}\sum_{j=1}^{|Y|}\delta(y_{j}\neq \hat{y_{j}}).
\end{equation}
\begin{equation}
\hat{y_j}={\arg\max}_{y}P(y|X)
\end{equation}

In the instance recognition subproblem, $\hat{Z}$ denotes the instance prediction. The cost function is:
\begin{equation}
Dist(Z,\hat{Z})=1-\frac{1}{K}\sum_{k'=0}^{K'}L(\widehat{R_{k'}},Z),
\end{equation}
where

\begin{equation}
  L(\widehat{R_{k'}},Z) = 
  \begin{cases}
    1, &\exists k\neq 0, \frac{\widehat{R_{k'}}\cap R_{k}}{\widehat{R_{k'}} \cup R_{k}}\geqslant thre\\
	0, &\text{otherwise}
  \end{cases}
\end{equation}

$\widehat{R_{k'}} \in \widehat{Z}$ denotes the instance segmentation prediction region and $R_{k} \in Z$ denotes the instance label region. $K'$ represents the total predicted region count. $thre$ is the threshold which is set to 0.5 in this algorithm. When the overlap ratio of the gland instance in a certain prediction region and labels is higher than the threshold, this region is considered an instance prediction by the algorithm.
Fig.~\ref{sub} shows the two gland instance segmentation subproblems.

Since the cost function of instance recognition is nondifferentiable, it cannot be trained with SGD. We hereby approximate instance recognition by edge detection and object detection. We generate edge labels $E$ and object labels $O$ through $Y$ and $Z$ to train edge detector and object detector, in which $E=\{e_{j},j=1,2,...,|X|\}, e_{j}\in \{0,1\}$ and $e_{j}$ equals 0 when all four nearest pixels (over, below, right and left) belong to the same instance. $O$ denotes the smallest bounding box for each gland instance.

\section{Related Work}
\label{related}
This section is a retrospective introduction about instance segmentation and gland instance segmentation.
\subsection{Instance segmentation}
Instance segmentation, a task distinguishing contour, location, class and the number of objects in an image, is attracting more and more attention from researchers in image processing and computer vision. As a complex problem can hardly be solved using traditional algorithms, a growing number of deep learning approaches have emerged to solve it. For example, SDS \cite{hariharan2014simultaneous} uses a framework that resembles R-CNN \cite{girshick2014rich} to extract features from both the bounding box of the region and the region foreground, and then classifies region proposals and refines the segmentation inside bounding boxes based on those extracted features. Hypercolumn \cite{hariharan2015hypercolumns} defines pixel features as a vector of activations of all CNN units above that pixel, and then classifies region proposals and refines region segmentation based on those feature vectors.  MNC \cite{dai2015instance} integrates three networks designed for detection, segmentation and classification respectively in a cascaded structure. Unlike SDS and Hypercolumn, MNC is capable of training in an end-to-end fashion, since MNC takes advantage of the Region Proposal Network (RPN) to generate region proposals. Similar to SDS and hypercolumn, MNC performs segmentation inside the proposal box as well. In contrast to the above methods, our method performs segmentation and instance recognition in a parallel manner.

\subsection{Gland instance segmentation} 
Gland morphology and structure can vary significantly, which poses a big challenge in gland instance segmentation. Researchers have come up with several methods to solve this problem \cite{chen2016dcan,jin2016object,sirinukunwattana2016gland,Kainz2015Semantic}. Previous works focus on detecting gland structure like nuclei and lumen. Sirinukunwattana \emph{et al.} \cite{sirinukunwattana2015stochastic} model every gland as a polygon in which the vertices are located at the nucleus. Cheikh \emph{et al.} \cite{Cheikh2016A} propose a mathematical morphology method to characterize the spatial distribution of nuclei in histological images. Nguyen \emph{et al.} \cite{nguyen2012prostate} use texture and structural features to classify the basic components of glands, and then segment gland instance based on prior knowledge of gland structure. These methods perform well in benign images but are comparatively unsatisfactory when used on malignant images, which has been the impetus for creating methods based on deep learning \cite{sirinukunwattana2015stochastic}. Li \emph{et al.} \cite{Li2016Gland} train a window-based binary classifier to segment glands using both CNN features and hand-crafted features. Kainz \emph{et al.} \cite{Kainz2015Semantic} train two separated networks to recognize glands and gland-separating structures respectively. In MICCAI 2015 gland segmentation challenge contest, some teams achieved impressive performance. DCAN \cite{chen2016dcan} is a multitask learning framework that combines a down-sampling path and an up-sampling path together. From the hierarchical layer, the framework is separated into two branches to generate contour information and segment objects. Team ExB \cite{sirinukunwattana2016gland} proposes a multipath convolutional neural network segmentation algorithm. Each path consists of different convolutional layers and is designed to capture different features. All paths are fused by two fully connected layers to integrate information. Team Freburg \cite{sirinukunwattana2016gland} utilizes an off-the-shelf deep convolutional neural network U-net \cite{ronneberger2015u}, and then performs post-processing of hole-filling and removes objects less than 100 pixels wide from the final results.

\subsection{Previous work}
An earlier conference version of our approach was presented in Xu \emph{et al.} \cite{xu2016gland}. Here we further illustrate that: (1) we explore another channel - object detection - in this paper, due to the edge detection and the object detection channels complementing each other; (2) ablation experiments are carried out to corroborate the effectiveness of the proposed algorithm; (3) based on the rotation invariance of histological images, a new data augmentation strategy is proposed that has proven to be effective; (4) this algorithm achieves state-of-the-art results on the dataset provided by the 2015 MICCAI Gland Segmentation Challenge Contest.

\section{Method}
\label{method}
\begin{figure}[!htbp]\centering
\includegraphics[width=0.45\textwidth]{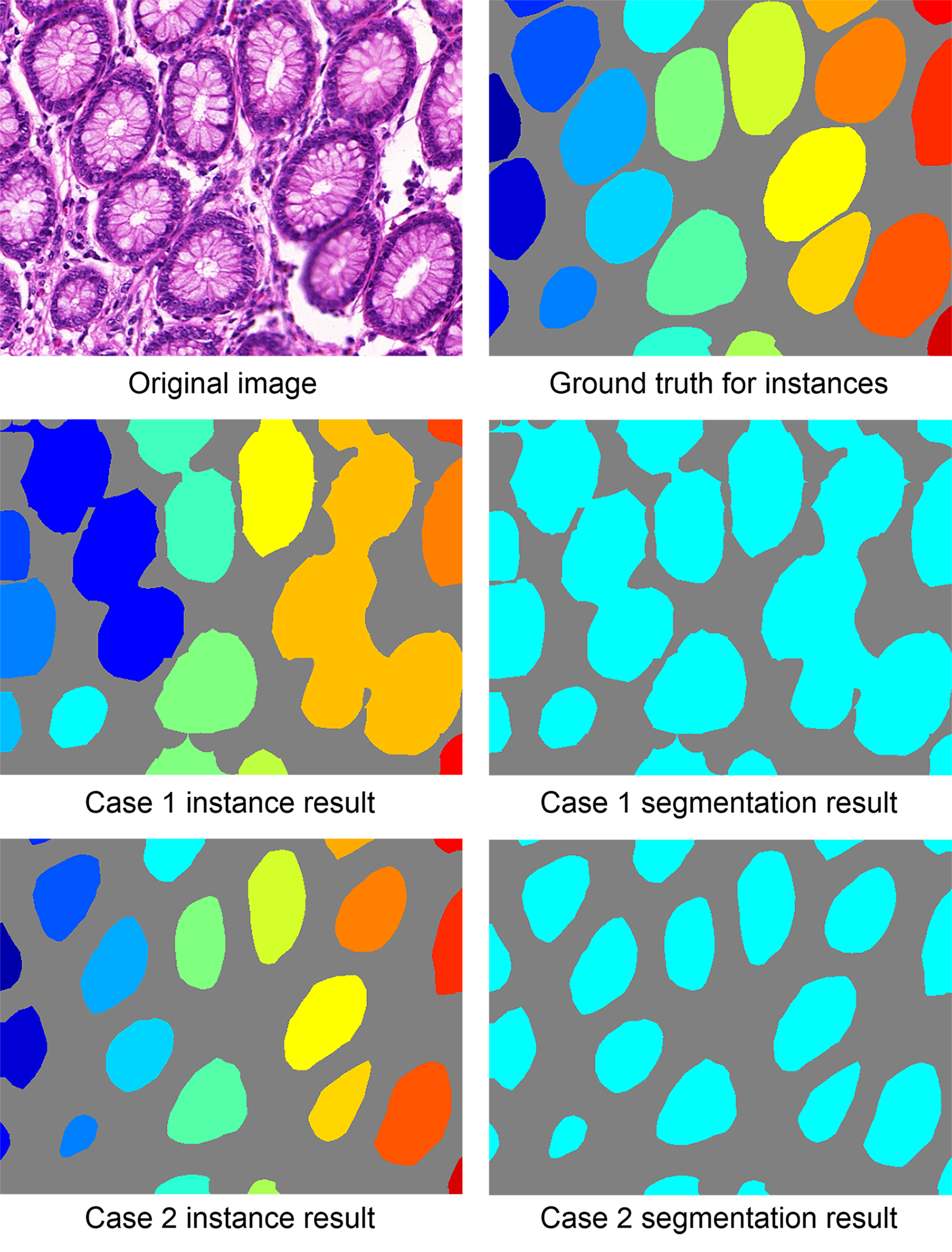}
	\caption{Two possible failures of gland instance segmentation. In the instance results and the ground truth images, different color regions represent different gland instances. Case 1 and Case 2 are two possible scenarios in which the algorithm fails to segment gland instances. In Case 1, glands are separated from the background but instances are not recognized. In Case 2, instances are labeled yet under the condition of many gland pixels being neglected.}
	\label{sce}
\end{figure}

There are two possible failures for gland instance segmentation. Since the gland-separating tissues are relatively few and similar to glands in coloration, it is very difficult for segmentation to rule out those pixels completely. Although it has little effect on segmentation, it is detrimental to the instance recognition process. Only one pixel that connects two glands can mislead the algorithm into recognizing that they belong to the same gland.
Another possible scenario is that algorithms designed to recognize instances separately may cause prediction areas to be smaller than the ground truth. In this case, the objects number and position may be accurate, but the segmentation performance is substandard. Those two scenarios are illustrated in Fig.~\ref{sce}.

\begin{figure*}[!t]
	\centering
	\includegraphics[width=0.9\textwidth]{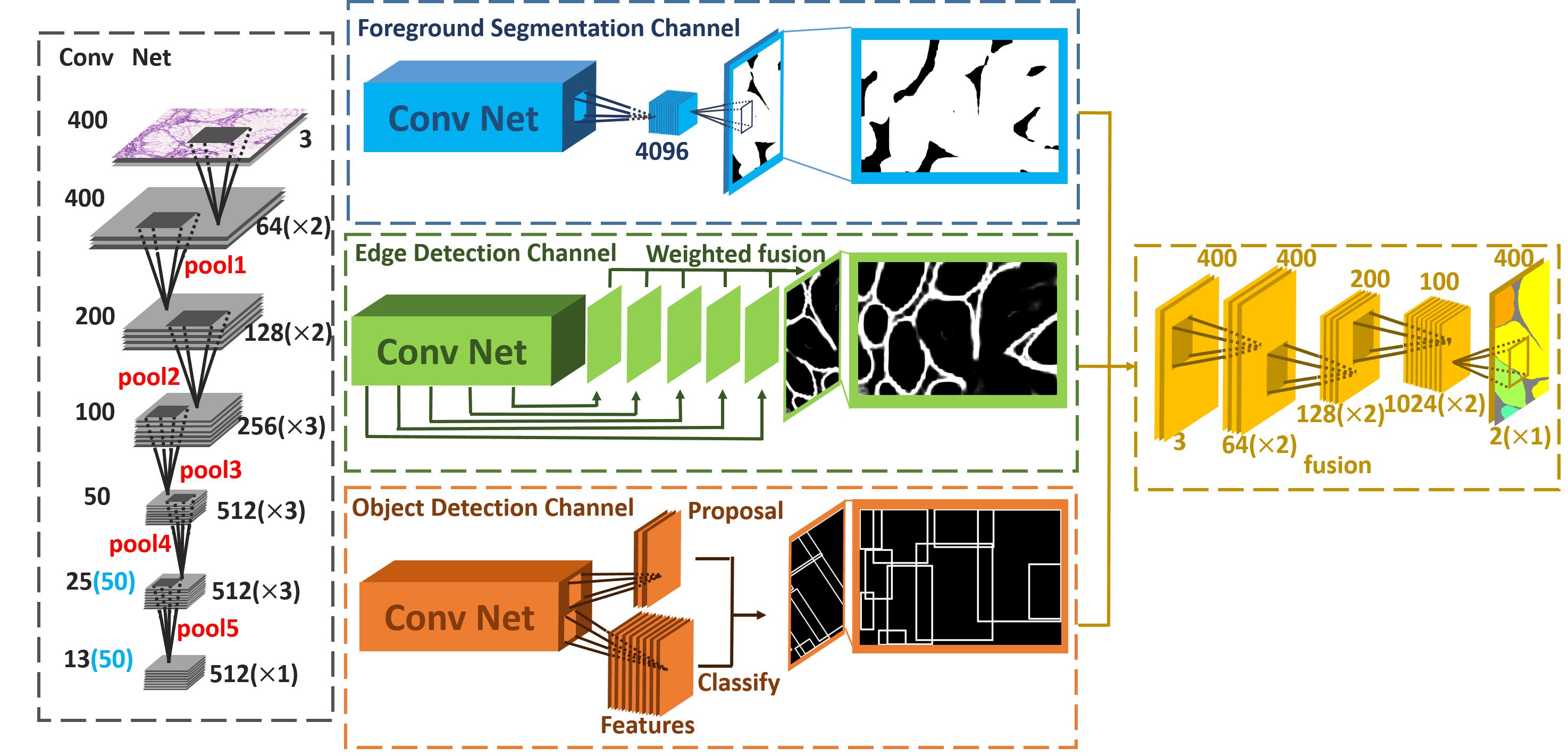}
	\caption{This illustrates the structure of this algorithm. For all the channels in this algorithm, FCN for the foreground segmentation channel, Faster R-CNN \cite{ren2015faster} for the object detection channel and HED the for edge detection channel, are all based on the VGG16 model, we present this classical five pooling structure in detail by ``Conv Net'' at the left side of the figure and represent it as a rectangular block named ``Conv Net''. Especially in foreground segmentation and object detection channels, arrows pointing from ``Conv Net'' denote the output of the ``Conv Net'', whereas in the edge detection channel they represent the output of deep supervisions. In the foreground segmentation channel, strides of the last two pooling layers of ``Conv Net'' are set as 1; dilated convolution is applied to convolution layers leading to the higher resolution of feature maps (as annotated in brackets in blue). In edge detection channel and object detection channel, the stride of pool4 and pool5 is 2. The `$\times2$' in brackets means that there are two convolutional layers.}
	\label{model2}
\end{figure*}

We propose a new multichannel algorithm to achieve gland segmentation and gland instance recognition simultaneously. Our algorithm consists of three channels and each of them is designed to undertake different responsibilities. In the proposed algorithm, we generate one kind of label of the input image for each channel. Fig.~\ref{model1} presents the flow chart of the proposed algorithm. One channel is designed to segment foreground pixels from background pixels. The other two channels are used to recognize instances. Aiming to determine which gland each foreground pixel belongs to, we utilize both object detection and edge detection to define spatial limits of every gland. The reason for choosing these two channels is based on the fact that information on contour and location contributes respectively and complimentarily to instance recognition and the joint effort will perform much better together than each one alone. Specifically, edge detection performs a little better than object detection in instance recognition, but edge detection fails to complete the task because of the aforementioned coalescence phenomenon of glands, which affects not only segmentation but edge detection as well. Gland detection may perform well for benign and well-shaped glands, but hardly detect the entire glands accurately for malignant ones. However, edge detection and object detection can compensate for each other's weaknesses and identify instances better. By integrating the information generated from different channels, our multichannel framework is capable of instance segmentation. A detailed depiction of our algorithm is presented in Fig ~\ref{model2}.

\subsection{Foreground Segmentation Channel}
The foreground segmentation channel distinguishes glands from the background.

The well-suited solutions to image labeling/segmentation in which each pixel is assigned a label from a pre-specified set are FCN family models \cite{long2015fully,xie15hed}. FCN replaces the fully-connected layer with a convolutional layer and upsamples the feature map to the same size as the original images through deconvolution thus an end-to-end training and prediction is guaranteed. Compared to the previous prevalent method,  sliding window \cite{sermanet2013overfeat,ciresan2012deep} in image segmentation, FCN is faster and simpler.  Usually, an FCN model can be regarded as the combination of a feature extractor and a pixel-wise predictor. A pixel-wise predictor predicts probability masks of segmented images. The feature extractor is able to abstract high-level features by down-sampling and convolution. Though useful high-level features are extracted, details of images sink in the process of max-pooling and strided convolution. Consequently, when objects are adjacent to each other, FCN may consider them as one. Applying FCN to segment images is a logical choice but instance segmentation is beyond the ability of FCN. It requires an algorithm to differentiate instances of the same class even when they are extremely close to each other. Even so, probability masks produced by FCN still offer valuable support in solving instance segmentation problems. 

To compensate for the resolution reduction of feature maps due to downsampling, FCN introduces skip architecture to combine deep semantic information and shallow appearance information. Nevertheless, Yu \emph{et al.} \cite{yu2015multi} propose the dilated convolution that empowers the network with a wider receptive field without downsampling. Less downsampling means less space-invariance brought by downsampling which is beneficial to increasing segmentation precision. 

Our foreground segmentation channel is a modified version of the FCN-32s \cite{long2015fully} of which the strides of pool4 and pool5 are 1 and subsequent convolution layers enlarge the receptive field with a dilated convolution. 

Given an input image $X$ and the parameter of the FCN network is denoted as ${w}_{s}$, thus the output of FCN is 
\begin{equation}
{P}_{s}\left(Y^{*}=k \mid X;w_{s}\right) = {\mu}_{k}\left(h_{s}\left(X,w_{s}\right)\right),
\end{equation}
where $\mu(\cdot)$ is the softmax function. $\mu_{k}(\cdot)$ is the output of the $k$th category and $h_{s}(\cdot)$ outputs the feature map of the hidden layer. In this case, there are two categories (foreground/glands and background), k=2. $Y^{*}$ is the segmentation prediction.

We train the foreground segmentation channel using softmax cross entropy loss.

\subsection{Edge Detection Channel}
The edge detection channel detects boundaries between glands.

To receive precise and clear boundaries, edges are crucial as proven by DCAN \cite{chen2016dcan}. The effectiveness of edges in our algorithm can be shown in two ways. First, the edge compensates for the information loss caused by max-pooling and strided convolution in FCN. As a result, contours become more precise and the morphology becomes more similar to the ground truth. Second, even if the location and the probability mask are confirmed, it is unavoidable that predicted pixel regions of adjacent objects are still connected. Edge, however, is able to differentiate between them. 
As expected, the synergy of regions, locations and edges achieves state-of-the-art results. The edge channel in our model is based on a Holistically-nested Edge Detector (HED) \cite{xie15hed}. It is a CNN-based solution towards edge detection. It learns hierarchically embedded multiscale edge fields to account for the low-, mid-, and high- level information of contours and object boundaries. In edge detection, pixels of labels are much less than pixels of backgrounds. The imbalance may decrease the convergence rate or even cause the network being unable to convergence. To solve the problem, deep supervision \cite{lee2015deeply} is deployed. In total, there are five side supervisions which are established before each down-sampling layer. 

We denote $w_{e}$ as the parameter of HED, thus the $m$th prediction of deep supervision is
\begin{equation}
P^{(m)}_{e}(E^{(m)*}=1 \mid X;w_{e})=\sigma(h^{(m)}_{e}(X,w_{e})).
\end{equation}
$\sigma(\cdot)$ denotes the sigmoid function - the output layer of HED. $h^{(m)}_{e}$ represents the output of the hidden layer relative to $m$th deep supervision and $E^{(m)*}$ denotes the $m$th side output prediction. The weighted sum of M outputs of deep supervision is the final result of this channel which is denoted as $E^*$, and the weighted coefficient is $\alpha$. 
\begin{equation}
P_{e}(E^*=1|X;w_e,\alpha)=\sigma(\sum_{m=1}^{M}\alpha^{(m)},h_{e}^{(m)}(X,w_e))
\end{equation}
This process is delivered through the convolutional layer. The back propagation enables the network to learn relative levels of importance of edge predictions under different scales.

We train the edge detection channel using sigmoid cross entropy loss.

\subsection{Object Detection Channel}
The object detection channel detects glands and their locations in the image. 

Object detection is helpful in counting and identifying the range of objects. According to some previous works on instance segmentation, such as MNC \cite{dai2015instance}, confirmation of the bounding-box is usually the first step in instance segmentation. After that, segmentation and other options are carried out within bounding boxes. Though this method is widely recognized, the loss of context information caused by the limited receptive field of bounding-box may exacerbate segmentation results. Consequently, we integrate location information into the fusion network instead of segmenting instances within bounding boxes.
To obtain location information, Faster R-CNN, a state-of-the-art object detection model, is conceived. Convolutional layers are applied to extract feature maps from images. After that, the Region Proposal Network (RPN) takes an arbitrary-sized feature map as input and produces a set of bounding-boxes with the probability of objects. Region proposals will be converted into regions of interest and classified to form the final object detection result. 

Filling is done in order to transform the bounding box prediction into a new formation that represents the number of bounding boxes that every pixel belongs to. The value of each pixel in regions covered by the bounding boxes equals the number of bounding boxes it belongs to. For example, if a pixel is in the overlapping area of three bounding boxes, the value of that pixel will be three. $w_{d}$ is denoted as the parameter of Faster R-CNN and $\phi$ represents the filling operation. The output of this channel is 
\begin{equation}
P_{d}\left(X,w_{d}\right) = \phi\left(h_{d}\left(X,w_{d}\right)\right).
\end{equation}
$h_{d}\left(\cdot\right)$ is the predicted coordinate of the bounding box.

We train the object detection channel using the same loss as in Faster R-CNN $\cite{ren2015faster}$: the sum of a classification loss and a regression loss.

\subsection{Fusing Multichannel}
Merely receiving the information of these three channels is not the ultimate purpose of our algorithm. As a result, a fusion algorithm is of great importance to maximize synergies of the three kinds of information - region, location and boundary cues. It is hard for an algorithm which is not learning-based to recognize the patterns of all this information. Naturally, a CNN based solution is the best choice.

After obtaining outputs of these three channels, a shallow seven-layer convolutional neural network is used to combine information and yield the final result. To reduce information loss and ensure a sufficiently large reception field, we again replace downsampling with dilated convolution.
The architecture of fusion network is designed by cross validation. We gradually increase the number layers and filters until the performance no longer improves. 

We denote $w_{f}$ as the parameter of this network and $h_{f}$ as the hidden layer. Thus the output of the network is
\begin{equation}
P\left(Y^{*}_{I}=k\mid P_{s},P_{d},P_{e};w_{f}\right)=\mu_{k}\left(h_{f}\left(P_{s},P_{d},P_{e},w_{f}\right)\right).
\end{equation}
As mentioned above, in this case, there are two categories, k=2.
$Y^{*}_{I}$ is the instance segmentation prediction.

We train the fusion network using softmax cross entropy loss.

\section{Experiment}
\label{exp}

\subsection{Dataset}
Our method is evaluated on the dataset provided by the MICCAI 2015 Gland Segmentation Challenge Contest \cite{sirinukunwattana2016gland}. The dataset consists of 165 labeled colorectal cancer histological images scanned by Zeiss MIRAX MIDI. The image resolution is approximately 0.62μm per pixel. Original images are in different sizes, while most of them are $775\times522$. 85 images belong to the training set and 80 are part of test sets (test set A contains 60 images and test set B contains 20 images). There are 37 benign sections and 48 malignant ones in the training set, 33 benign sections and 27 malignant ones in testing set A and 4 benign sections and 16 malignant ones in testing set B.
\subsection{Data augmentation and Preprocessing}
We first preprocess data by performing per channel zero mean. The next step is to generate edge labels from region labels and perform dilation on edge labels afterwards. A bounding box for a gland is the smallest rectangle that can encircle the gland. Bounding box ground truth ($x^k_{min}$,$x^k_{max}$,$y^k_{min}$,$y^k_{max}$) can be generated from segmentation label, in which, $x^k_{min}=\min({P_x|P\in R_k})$, $x^k_{max}=\max({P_x|P\in R_k})$, $y^k_{min}=\min({y_x|P\in R_k})$, and $y^k_{max}=\max({P_y|P\in R_k})$. $R_k$ is the $k$th region of the instance ground truth and $P$ denotes a pixel point in $R_k$. $P_x$ and $P_y$ represent the X-coordinate and Y-coordinate of $P$. Whether a pixel is an edge or not is decided by its four nearest pixels (over, below, right and left) in the region label. If all four pixels in the region label belong to the foreground or in the background, this pixel does not belong to any edge. To enhance performance and combat overfitting, copious amounts of training data are needed. Given the circumstance of the absence of a large dataset, data augmentation is essential before training. Two strategies for data augmentation have been carried out and the improvement of results is strong enough evidence to prove the efficiency of data augmentation. In Strategy \uppercase\expandafter{\romannumeral1}, horizontal flipping and rotation operation ($0^\circ$, $90^\circ$, $180^\circ$, $270^\circ$) are used in training images. Besides operations in Strategy \uppercase\expandafter{\romannumeral1}, Strategy \uppercase\expandafter{\romannumeral2} also includes elastic transformation, such as pin cushion transformation and barrel transformation. Deformation of original images is beneficial to increasing robustness and the promotion of the final result. Since the fully-connected layer is replaced by convolutional layer, FCN takes arbitrary size images as testing inputs. After data augmentation, a $400 \times 400$ region is randomly cropped from the original image as input.

\subsection{Hyperparameter}
CAFFE $\cite{jia2014caffe}$ is used in our experiments. Experiments are carried out on K40 GPU and the CUDA edition is 7.0. The weight decay is 0.002, the momentum is 0.9. While training the foreground labeling/segmentation channel of the network, the learning rate is $10^{−3}$ and the parameters are initialized by pre-trained FCN32s model $\cite{long2015fully}$, while the edge detection channel is trained under the learning rate of $10^{−9}$ and the Xavier initialization is performed. object detection channel is trained under the learning rate of $10^{-3}$ and initialized by pretrained Faster R-CNN model. Fusion is learned under the learning rate of $10^{-3}$ and initialized by Xavier initialization.

\subsection{Evaluation}
The evaluation method is the same as the competition requires. Three indicators are used to evaluate the performance on test A and test B. Indicators assess detection results, segmentation performance and shape similarity respectively. The final score is the summation of six rankings and the smaller the better. Since image amounts of test A and test B have a significant difference in quantity, we not only calculate the rank sum as the host of MICCAI 2015 Gland Segmentation Challenge Contest demands, but we also list the weighted rank sum. We calculate the weighted average of three evaluation criteria on test set A and test set B. Since the images in test A account for 3/4 of the test set and images in test B account for 1/4, the weighted rank sum is calculated as:
\begin{equation}
Weighted RS=3/4\sum test A Rank+1/4\sum test B Rank.
\end{equation}
The evaluation program is given by the MICCAI 2015 Gland Segmentation Challenge Contest \cite{sirinukunwattana2016gland}.
The first criterion is the $F_{1}$ score, which reflects gland detection accuracy. The segmented glandular object of True Positive (TP) is the object that shares more than 50\% of areas with the ground truth. Otherwise, the segmented area will be determined as a False Positive (FP). Objects of ground truth without corresponding prediction are considered as False Negatives (FN).
\begin{equation}
F1\quad Score = \frac{2\cdot Precision\cdot Recall}{Precision + Recall}
\end{equation}
\begin{equation}
Precision = \frac{TP}{TP+FP}
\end{equation}
\begin{equation}
Recall=\frac{TP}{TP+FN}
\end{equation}

\begin{table*}[!hbtp]
	\normalsize
	\caption{Performance in Comparison to Other Methods}
	\resizebox{\textwidth}{!}{ 
		\begin{tabular}{c|c|c|c|c|c|c|c|c|c|c|c|c|c|c}
			\hline\hline
			\multirow{3}{*}{Method}&
			\multicolumn{4}{c|}{$F_{1}$ Score}&
			\multicolumn{4}{c|}{ObjectDice}&
			\multicolumn{4}{c|}{ObjectHausdorff}&
			\multirow{3}{*}{RS\footnotemark[1]}&
			\multirow{3}{*}{WRS\footnotemark[2]}\\
			\cline{2-13}
			&\multicolumn{2}{c|}{Part A}&
			\multicolumn{2}{c|}{Part B}&
			\multicolumn{2}{c|}{Part A}&
			\multicolumn{2}{c|}{Part B}&
			\multicolumn{2}{c|}{Part A}&
			\multicolumn{2}{c|}{Part B}& \\
			\cline{2-13}
			& Score & Rank &  Score & Rank & Score & Rank & Score & Rank & Score & Rank & Score & Rank & \\
			\hline
			FCN \cite{long2015fully} & 0.788 & 11 & 0.764 & 4 & 0.813 & 11 & 0.796 & 4 & 95.054 & 11 & 146.2478 & 4 & 45 & 27.75 \\
			dilated FCN \cite{chen2016deeplab} & 0.854 & 9 & 0.798 & 2 & 0.879 & 6 & 0.825 & 2 & 62.216 & 9 &  118.734 & 2 & 30 & 19.5 \\
			\hline
			\textbf{\cellcolor[rgb]{.9,.9,.9}Ours}  & \cellcolor[rgb]{.9,.9,.9}0.893 & \cellcolor[rgb]{.9,.9,.9}3 & \textbf{\cellcolor[rgb]{.9,.9,.9}0.843} & \textbf{\cellcolor[rgb]{.9,.9,.9}1} & \textbf{\cellcolor[rgb]{.9,.9,.9}0.908} & \textbf{\cellcolor[rgb]{.9,.9,.9}1} & \textbf{\cellcolor[rgb]{.9,.9,.9}0.833} & \textbf{\cellcolor[rgb]{.9,.9,.9}1} & \textbf{\cellcolor[rgb]{.9,.9,.9}44.129} & \textbf{\cellcolor[rgb]{.9,.9,.9}1} & \textbf{\cellcolor[rgb]{.9,.9,.9}116.821} & \textbf{\cellcolor[rgb]{.9,.9,.9}1} & \cellcolor[rgb]{.9,.9,.9}8 & \cellcolor[rgb]{.9,.9,.9}4.5\\
			CUMedVision2 \cite{chen2016dcan} & \textbf{0.912} & \textbf{1} & 0.716 & 6 & 0.897 & 2 & 0.781 & 8 & 45.418 & 2 & 160.347 & 9 & 28 & 9.5\\
			ExB3 \cite{sirinukunwattana2016gland} & 0.896 & 2 & 0.719 & 5 & 0.886 & 3 & 0.765 & 9 & 57.350 & 6 & 159.873 & 8 & 33 & 13.75\\
			ExB2 \cite{sirinukunwattana2016gland} & 0.892 & 4 & 0.686 & 9 & 0.884 & 4 & 0.754 & 10 & 54.785 & 3 & 187.442 & 11 & 41 & 15.75\\
			ExB1 \cite{sirinukunwattana2016gland} & 0.891 & 5 & 0.703 & 7 & 0.882 & 5 & 0.786 & 5 & 57.413 & 7 & 145.575 & 3 & 32 & 16.5\\
			Frerburg2 \cite{ronneberger2015u} & 0.870 & 6 & 0.695 & 8 & 0.876 & 7 & 0.786 & 6 & 57.093 & 4 & 148.463 & 6 & 37 & 17.75\\
			Frerburg1 \cite{ronneberger2015u} & 0.834 & 10 & 0.605 & 11 & 0.875 & 8 & 0.783 & 7 & 57.194 & 5 & 146.607 & 5 & 46 & 23\\
			CUMedVision1 \cite{chen2016dcan} & 0.868 & 7 & 0.769 & 3 & 0.867 & 10 & 0.800 & 3 & 74.596 & 10 & 153.646 & 7 & 40 & 23.5\\
			CVIP Dundee & 0.863 & 8 & 0.633 & 10 & 0.870 & 9 & 0.715 & 11 & 58.339 & 8 & 209.048 & 13 & 59 & 27.25\\
			LIB & 0.777 & 12 & 0.306 & 14 & 0.781 & 12 & 0.617 & 13 & 112.706 & 13 & 190.447 & 12 & 76 & 37.5\\
			CVML & 0.652 & 13 & 0.541 & 12 & 0.644 & 14 & 0.654 & 12 & 155.433 & 14 & 176.244 & 10 & 75 & 39.25\\
			vision4GlaS & 0.635 & 14 & 0.527 & 13 & 0.737 & 13 & 0.610 & 14 & 107.491 & 12 & 210.105 & 14 & 80 & 39.5\\
			\hline\hline
		\end{tabular}
	}
		\scriptsize
		\\
	\footnote -RS is the abbreviation for rank sum.
	\\
	\footnote -WRS is the abbreviation for weighted rank sum.
	\label{rank}
\end{table*}

Dice is the second criterion for evaluating segmentation performance. The dice index of the whole image is
\begin{equation}
D(G,S)=\frac{2(\mid G\cap S\mid)}{\mid G\mid +\mid S\mid},
\end{equation}
of which $G$ represents the ground truth and $S$ is the segmented result. Unfortunately, it is not able to differentiate instances of the same class. Further, we denote $G$ as a set of all ground truth objects and $S$ as a set of all segmented objects. $S_i$ denotes the $i$th segmented object in an image and $G_i$ denotes a ground truth object that maximally overlaps $S_i$ in the image. $\widetilde{G}_{i}$ denotes the $i$th ground truth object in and image and $\widetilde{S}_{i}$ denotes a segmented object that maximally overlaps in the image. As a result, an object-level dice score is employed to evaluate segmentation results. The definition is as follows:
\begin{equation}
D_{object}(G,S)=1/2\left[\sum_{i=1}^{n_{S}}w_{i}D(G_{i},S_{i})+\sum_{i=1}^{n_{G}}\widetilde{w}_{i}D(\widetilde{G}_{i},\widetilde{S}_{i})\right],
\end{equation}
\begin{equation}
w_{i}=\frac{\mid S_{i}\mid}{\sum_{j=1}^{n_{S}}\mid S_{j}\mid},
\end{equation}
\begin{equation}
\widetilde{w}_{i}=\frac{\mid \widetilde{G}_{i}\mid}{\sum_{j=1}^{n_{G}}\mid \widetilde{G}_{j} \mid}.
\end{equation}
$n_{S}$ and $n_{G}$ are the numbers of instances in the segmented results and the ground truth.

Shape similarity reflects the performance on morphology likelihood which plays a significant role in gland instance segmentation. Hausdorff distance is exploited to evaluate shape similarity. To assess glands respectively, the index of Hausdorff distance deforms from the original formation:
\begin{equation}
H(G,S)=\mathrm{max}\left\{\underset{x\epsilon G}{sup}  \underset{y\epsilon S}{inf}\left\|x-y\right\|,\underset{y\epsilon S}{sup}  \underset{x\epsilon G}{inf}\left\|x-y\right\|\right\},
\end{equation}
to the object-level formation:
\begin{equation}
H_{object}(S,G)=1/2\left[\sum_{i=1}^{n_{s}}w_{i}H(G_{i},S_{i})+\sum_{i=1}^{n_{G}}\widetilde{w}_{i}H(\widetilde{G}_{i},\widetilde{S}_{i})\right],
\end{equation}
where
\begin{equation}
w_{i}=\frac{|S_{i}|}{\sum_{j=1}^{n_{S}}|S_{j}|},
\end{equation}
\begin{equation}
\widetilde{w}_{i}=\frac{|\widetilde{G}_{i}|}{\sum_{j=1}^{n_{G}}|\widetilde{G}_{j}|}.
\end{equation}
Similar to the object-level dice, index $n_{S}$ and $n_{G}$ represent instances of segmented objects and the ground truth.

\subsection{Result and Discussion}
Table~\ref{rank} lists results of our proposed algorithm, FCN, dilated FCN and other participants on datasets provided by the MICCAI 2015 Gland Segmentation Challenge Contest. 

In the table, RS and WRS denote rank sum and weighted rank sum respectively.
We rearrange the scores and ranks in this table. Our method outranks FCN, dilated FCN and other participants based on both rank sum and weighted rank sum.

\begin{figure*}[ht]
	\centering
	\includegraphics[width=0.93\textwidth]{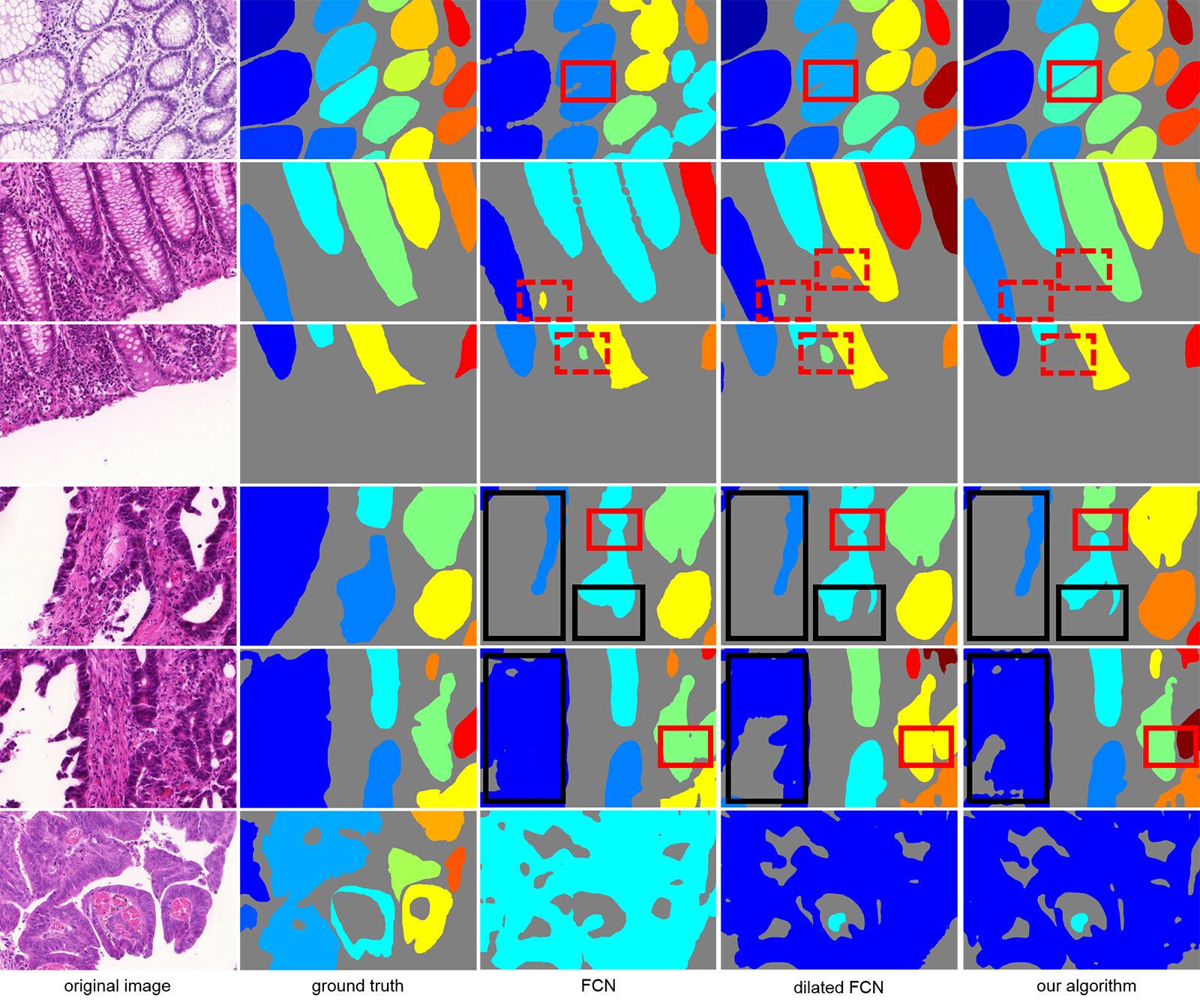}
	\caption{From left to right: original image, ground truth, results of FCN \cite{long2015fully}, FCN with dilated convolution and the proposed algorithm. Compared to FCN and dilated FCN, most adjacent glandular structures are separated (as shown inside the red solid boxes) which indicates that our algorithm accomplishes instance segmentation. Besides, our algorithm is able to correctly judge the small isolated area as non-gland area (as shown inside the red dotted boxes). However, a few glands that are broken apart escape the detection of our model (as shown inside  the black boxes). The bad performance in the last row is due to the fact that in most samples the white area is recognized as cytoplasm whereas in this sample, the white area is the background.}
	\label{result}
\end{figure*}

Compared to FCN and dilated FCN, our algorithm obtains better scores which is convincing evidence that our work is more effective in solving instance segmentation problems in histological images. Though dilated FCN performs better than FCN as the dilated convolution process has less pooling and covers larger receptive fields, our algorithm combines region, location and edge information to achieve higher scores in the dataset. The reason our algorithm ranks higher is because most adjacent glandular structures have been separated, which is more beneficial to meet the evaluation index of instance segmentation, whereas in FCN and dilated FCN they are not. Comparison results are illustrated in Fig.~\ref{result}.

Ranks of test A are generally higher than test B due to the inconsistency of data distribution. In test A, most images are normal ones whereas test B contains a majority of cancerous images which are more complicated in shape and larger in size. Hence, a larger receptive field is required in order to detect cancerous glands. However, before we exploit dilated convolution, the downsampling layer not only gives the network a larger receptive field but also makes the resolution of the feature map decrease, thus it deteriorates the segmentation results. Dilated convolution empowers the convolutional neural network with a larger receptive field with fewer downsampling layers. Our multichannel algorithm enhances performance based on the dilated FCN by adding two channels - edge detection and object detection.

Since the differences between background and foreground in histopathological images are small (3th row of Fig.~\ref{result}), FCN and dilated FCN sometimes predict the background pixel as gland, raising the false positive rate. The multichannel algorithm abates the false positive by adding pixel context while predicting object location. 

Compared to CUMedVision1 \cite{chen2016dcan}, CUMedVision2 \cite{chen2016dcan} adds edge information which improves the results of test A but those of test B deteriorate. Our method improves results of test A and test B after combining edge and location context. 

\begin{figure*}[!ht]
	\centering
	\includegraphics[width=0.93\textwidth]{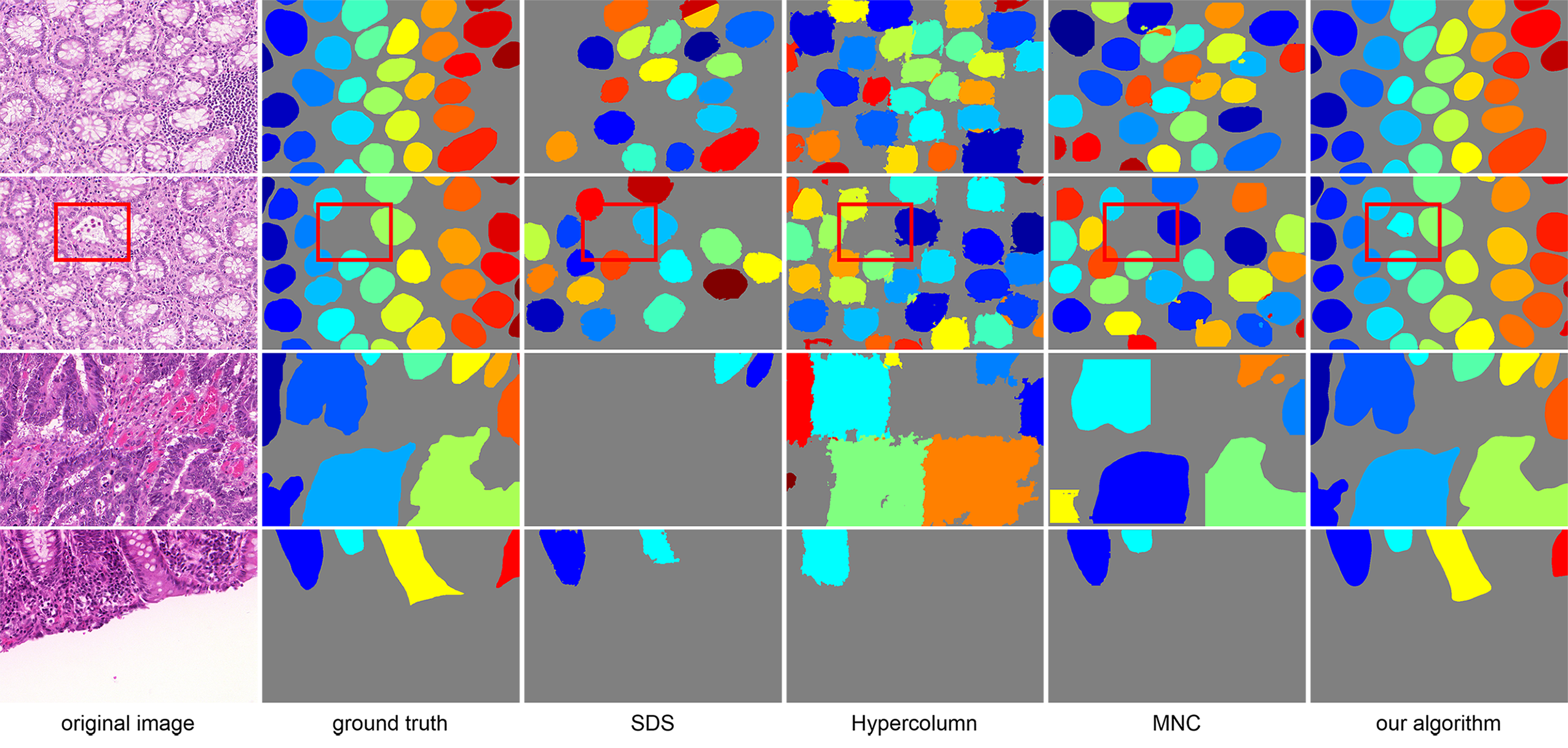}
	\caption{From left to right: original image, ground truth, results of SDS \cite{hariharan2014simultaneous}, Hypercolumn \cite{hariharan2015hypercolumns}, MNC \cite{dai2015instance} and the proposed algorithm. Different color regions represent different gland instances. SDS, Hypercolumn and MNC all perform masking inside bounding boxes produced by object detection, which causes the coarse boundary of gland instances and even neglects some glands. In the second column, one gland instance is missed by manual labeling but our algorithm successfully detects its location and segments it with relatively complete shape, yet SDS, Hypercolumn and MNC fail to detect this gland (as shown inside of the red boxes). }
	\label{figins}
\end{figure*}

However, white regions in gland histopathological images are of two kinds: 1) cytoplasm; and 2) no cell or tissue (background). The difference between these two is that cytoplasm usually appears surrounded by nuclei or other stained tissue. In the image of the last row in Fig.~\ref{result}, glands encircle some white regions with no existence of  cell or tissue causing the algorithm to mistake them for cytoplasm. As for images of the 4th and 5th row in Fig.~\ref{result}, glands are split when cutting images, which is the reason that cytoplasm is mistaken for background.

\begin{table*}[!h]
	\centering
	\tiny
	\caption{Comparison with instance segmentation methods}
	\resizebox{0.8\textwidth}{!}{
		\begin{tabular}{c|c|c|c|c|c|c}
			\hline\hline
			\multirow{2}{*}{Method} & 
			\multicolumn{2}{c|}{$F_{1}$ Score} &
			\multicolumn{2}{c|}{ObjectDice} &
			\multicolumn{2}{c}{ObjectHausdorff} \\
			\cline{2-7}
			& Part A & Part B & Part A & Part B & Part A & Part B\\
			\hline
			HyperColumn \cite{hariharan2015hypercolumns} & 0.852 & 0.691 & 0.742 & 0.653 & 119.441 & 190.384\\
			MNC \cite{dai2015instance} & 0.856 & 0.701 & 0.793 & 0.705 & 85.208 & 190.323\\
			SDS \cite{hariharan2014simultaneous} & 0.545 & 0.322 & 0.647 & 0.495 & 116.833 & 229.853\\
			BOX-$>$dilated FCN \cite{chen2016deeplab}+EDGE3 & 0.807 & 0.700 & 0.790 & 0.696 & 114.230 & 197.360\\
			\cellcolor[rgb]{.9,.9,.9}OURS & \cellcolor[rgb]{.9,.9,.9}\textbf{0.893} & \cellcolor[rgb]{.9,.9,.9}\textbf{0.843} & \cellcolor[rgb]{.9,.9,.9}\textbf{0.908} & \cellcolor[rgb]{.9,.9,.9}\textbf{0.833} & \cellcolor[rgb]{.9,.9,.9}\textbf{44.129} & \cellcolor[rgb]{.9,.9,.9}\textbf{116.821}\\
			\hline\hline
		\end{tabular}}
		\label{ins}
	\end{table*}

\textbf{Comparison with instance segmentation methods}
Currently, methods suitable for instance segmentation of natural scene images predict instances based on detection or proposal, such as SDS \cite{hariharan2014simultaneous}, Hypercolumn \cite{hariharan2015hypercolumns} and MNC \cite{dai2015instance}. One problem with this logic is its dependence on the precision of detection or proposal. If the object or a certain pixel of an object escapes the detection, it will evade the subsequent segmentation as well. Besides, the segmentation being restricted to a certain bounding box will have little access to context information hence it impacts the result. Under the condition of bounding boxes overlapping one another, which instance the pixel in the overlapping region belongs to cannot be determined. The overlapping area falls into the category of the nearest gland in our experiment. The experiment results are presented in Fig.~\ref{figins}.

To further demonstrate the defect of the cascade architecture, we design a baseline experiment. We first perform gland detection and then segment gland instances inside bounding boxes. There is a shallow network (same as the fusion network) combining foreground segmentation and edge detection information to generate the final result. Configurations of all experiments are set the same as our method. Results are shown in Table~\ref{ins} and less effective than the proposed algorithm.

\subsection{Ablation Experiment}
\subsubsection{Data Augmentation Strategy}
Data augmentation contributes to performance enhancement and overfitting elimination. We observe through experiments that adequate transformation of gland images is beneficial to training. This is because glands naturally form in various shapes and cancerous glands are more different in morphology. Here we evaluate the effect on results of the foreground segmentation channel using Strategy \uppercase\expandafter{\romannumeral1} and Strategy \uppercase\expandafter{\romannumeral2} (as shown in Table~\ref{data}).
	
\begin{table}[h]
\centering
\Large
\caption{Data Augmentation Strategy comparison}
\resizebox{0.49\textwidth}{!}{
	\begin{tabular}{c|c|c|c|c|c|c|c}
		\hline\hline
		\multirow{2}{*}{Strategy} &
		\multirow{2}{*}{Method} & 
		\multicolumn{2}{c|}{$F_{1}$ Score} &
		\multicolumn{2}{c|}{ObjectDice} &
		\multicolumn{2}{c}{ObjectHausdorff} \\
		\cline{3-8}
		& & Part A & Part B & Part A & Part B & Part A & Part B\\
		\hline
		\multirow{2}{*}{Strategy \uppercase\expandafter{\romannumeral1}} &
		FCN \cite{long2015fully} & 0.709 & 0.708 & 0.748 & 0.779 & 129.941 & 159.639\\
		\cline{2-8}
		& dilated FCN \cite{chen2016deeplab} & 0.820 & 0.749 & 0.843 & 0.811 & 79.768 & 131.639\\
		\hline
		\multirow{2}{*}{Strategy \uppercase\expandafter{\romannumeral2}} & 
		FCN \cite{long2015fully} & 0.788 & 0.764 & 0.813 & 0.796 & 95.054 & 146.248\\
		\cline{2-8}
		& dilated FCN \cite{chen2016deeplab} & \textbf{0.854} & \textbf{0.798} & \textbf{0.879} & \textbf{0.825} & \textbf{62.216} & \textbf{118.734}\\
		\hline\hline
    \end{tabular}}
    \label{data}
\end{table}

\subsubsection{Plausibility of Channels}
In convolutional neural networks, the main purpose of downsampling is to enlarge the receptive field, but this comes at a cost of decreased resolution and information loss of original data. Feature maps with low resolution increase the difficulty of upsample layer training. The representational ability of feature maps is reduced after upsampling and further leads to inferior segmentation results. Another drawback of downsampling is the space invariance it introduces whereas segmentation is space sensitive. The inconsistence between downsampling and image segmentation is obvious. Dilated convolution empowers the convolutional neural network with larger receptive field with less downsampling layers.
		
The comparison between segmentation performances of FCN with and without dilated convolution shows its effectiveness in enhancing segmentation precision. The foreground segmentation channel with dilated convolution improves the performance of the multichannel algorithm. So does the fusion stage with dilated convolution.
		
Pixels belonging to the edge occupy an extremely small proportion of the whole image. The imbalance between edge and non-edge poses a significant barrier to network training that the network may not convergent. Edge dilation can alleviate the imbalance and improve edge detection precision. 
		
To prove that these three channels truly improve instance segmentation performance, we conduct the following two baseline experiments: a) we launch a foreground segmentation channel and an edge detection channel; b) we launch a foreground segmentation channel and an object detection channel. The results favor the three-channel algorithm.	
Results from the experiments mentioned above are presented in Table~\ref{channel}.

\begin{table}[ht]
\Large
	\caption{Plausibility of Channels. }

	\resizebox{0.49\textwidth}{!}{
		\begin{tabular}{c|c|c|c|c|c|c}
			\hline\hline
			\multirow{2}{*}{Method} & 
			\multicolumn{2}{c|}{$F_{1}$ Score} &
			\multicolumn{2}{c|}{ObjectDice} &
			\multicolumn{2}{c}{ObjectHausdorff} \\
			\cline{2-7}
			& Part A & Part B & Part A & Part B & Part A & Part B\\
			\hline
			MC: FCN + EDGE1 + BOX & 0.863 & 0.784 & 0.884 & 0.833 & 57.519 & 108.825\\
			MC: FCN + EDGE3 + BOX & 0.886 & 0.795 & 0.901 & 0.840 & 49.578 & 100.681\\
			MC: dilated FCN + EDGE3 + BOX & 0.890 & 0.816 & \textbf{0.905} & 0.841 & 47.081 & 107.413\\
			\hline
			\hline
			DMC: FCN + EDGE3 + BOX & \textbf{0.893} & 0.803 & 0.903 & \textbf{0.846} & 47.510 & \textbf{97.440}\\
			DMC: dilated FCN + EDGE3 + BOX & \textbf{0.893} & \textbf{0.843} & 0.908 & 0.833 & \textbf{44.129} & 116.821\\
			DMC: dilated FCN + EDGE1 + BOX & 0.876 & 0.824 & 0.894 & 0.826 & 50.028 & 123.881\\
			\hline
			\hline
			DMC: dilated FCN + BOX & 0.876 & 0.815 & 0.893 & 0.808 & 50.823 & 132.816\\
			DMC: dilated FCN + EDGE3 & 0.874 & 0.816 & 0.904 & 0.832 & 46.307 & 109.174\\
			\hline\hline
		\end{tabular}}
		
	\footnotesize We denote DMC as the fusion network with dilated convolution \cite{chen2016deeplab} and MC as the fusion network without dilated convolution. EDGE1 represents that edge label are not dilated whereas EDGE3 represents that edge label are dilated by a disk filter with radius of 3. BOX indicates that the method includes object detection \cite{ren2015faster}. FCN \cite{long2015fully} and dilated FCN \cite{chen2016deeplab} indicates that the method includes foreground segmentation. 
		\label{channel}		
		\end{table}
			
\section{Conclusion}
\label{con}
We propose a new algorithm called deep multichannel neural networks. The proposed algorithm exploits features of edge, region and location in a multichannel manner to generate instance segmentation. We observe state-of-the-art results on the dataset from the MICCAI 2015 Gland Segmentation Challenge. A series of baseline experiments are conducted to prove the superiority of this method.			

In future work, this algorithm can be expanded to instance segmentation of other medical images.

\section*{Acknowledgment}
We thank the MICCAI 2015 Gland Segmentation Challenge for providing dataset. We thank Zhuowen Tu for all the help.
			
\ifCLASSOPTIONcaptionsoff
  \newpage
\fi

\bibliographystyle{IEEEtran}
\bibliography{IEEEabrv,./TBME}

\begin{thebibliography}{10}
\providecommand{\url}[1]{#1}
\csname url@samestyle\endcsname
\providecommand{\newblock}{\relax}
\providecommand{\bibinfo}[2]{#2}
\providecommand{\BIBentrySTDinterwordspacing}{\spaceskip=0pt\relax}
\providecommand{\BIBentryALTinterwordstretchfactor}{4}
\providecommand{\BIBentryALTinterwordspacing}{\spaceskip=\fontdimen2\font plus
\BIBentryALTinterwordstretchfactor\fontdimen3\font minus
  \fontdimen4\font\relax}
\providecommand{\BIBforeignlanguage}[2]{{%
\expandafter\ifx\csname l@#1\endcsname\relax
\typeout{** WARNING: IEEEtran.bst: No hyphenation pattern has been}%
\typeout{** loaded for the language `#1'. Using the pattern for}%
\typeout{** the default language instead.}%
\else
\language=\csname l@#1\endcsname
\fi
#2}}
\providecommand{\BIBdecl}{\relax}
\BIBdecl

\bibitem{travis2011international}
W.~D. Travis \emph{et~al.}, ``International association for the study of lung
  cancer/american thoracic society/european respiratory society international
  multidisciplinary classification of lung adenocarcinoma,'' \emph{Journal of
  Thoracic Oncology}, vol.~6, no.~2, pp. 244--285, 2011.

\bibitem{nguyen2012structure}
K.~Nguyen, A.~Sarkar, and A.~K. Jain, ``Structure and context in prostatic
  gland segmentation and classification,'' in \emph{MICCAI}.\hskip 1em plus
  0.5em minus 0.4em\relax Springer, 2012, pp. 115--123.

\bibitem{al2010improved}
Y.~Al-Kofahi \emph{et~al.}, ``Improved automatic detection and segmentation of
  cell nuclei in histopathology images,'' \emph{IEEE Transactions on Biomedical
  Engineering}, vol.~57, no.~4, pp. 841--852, 2010.

\bibitem{veta2013automatic}
M.~Veta \emph{et~al.}, ``Automatic nuclei segmentation in h\&e stained breast
  cancer histopathology images,'' \emph{PloS one}, vol.~8, no.~7, p. e70221,
  2013.

\bibitem{dimopoulos2014accurate}
S.~Dimopoulos \emph{et~al.}, ``Accurate cell segmentation in microscopy images
  using membrane patterns,'' \emph{Bioinformatics}, vol.~30, no.~18, pp.
  2644--2651, 2014.

\bibitem{naik2007gland}
S.~Naik \emph{et~al.}, ``Gland segmentation and computerized gleason grading of
  prostate histology by integrating low-, high-level and domain specific
  information,'' in \emph{MIAAB workshop}, 2007, pp. 1--8.

\bibitem{nguyen2010automated}
K.~Nguyen, A.~K. Jain, and R.~L. Allen, ``Automated gland segmentation and
  classification for gleason grading of prostate tissue images,'' in
  \emph{ICPR}, 2010, pp. 1497--1500.

\bibitem{naik2008automated}
S.~Naik \emph{et~al.}, ``Automated gland and nuclei segmentation for grading of
  prostate and breast cancer histopathology,'' in \emph{ISBI}, 2008, pp.
  284--287.

\bibitem{paul2016gland}
A.~Paul and D.~P. Mukherjee, ``Gland segmentation from histology images using
  informative morphological scale space,'' in \emph{ICIP}, 2016, pp.
  4121--4125.

\bibitem{egger2013pcg}
J.~Egger, ``Pcg-cut: graph driven segmentation of the prostate central gland,''
  \emph{PloS one}, vol.~8, no.~10, p. e76645, 2013.

\bibitem{tosun2011graph}
A.~B. Tosun and C.~Gunduz-Demir, ``Graph run-length matrices for
  histopathological image segmentation,'' \emph{IEEE Trans. Medical Imaging},
  vol.~30, no.~3, pp. 721--732, 2011.

\bibitem{sirinukunwattana2016gland}
K.~Sirinukunwattana \emph{et~al.}, ``Gland segmentation in colon histology
  images: The glas challenge contest,'' \emph{Medical Image Analysis}, vol.~35,
  pp. 489--502, 2016.

\bibitem{fleming2012colorectal}
M.~Fleming \emph{et~al.}, ``Colorectal carcinoma: pathologic aspects,''
  \emph{Journal of gastrointestinal oncology}, vol.~3, no.~3, pp. 153--173,
  2012.

\bibitem{hariharan2014simultaneous}
B.~Hariharan \emph{et~al.}, ``Simultaneous detection and segmentation,'' in
  \emph{ECCV}, 2014, pp. 297--312.

\bibitem{hariharan2015hypercolumns}
------, ``Hypercolumns for object segmentation and fine-grained localization,''
  in \emph{CVPR}, 2015, pp. 447--456.

\bibitem{dai2015instance}
J.~Dai, K.~He, and J.~Sun, ``Instance-aware semantic segmentation via
  multi-task network cascades,'' in \emph{CVPR}, 2016, pp. 3150--3158.

\bibitem{krizhevsky2012imagenet}
A.~Krizhevsky, I.~Sutskever, and G.~E. Hinton, ``Imagenet classification with
  deep convolutional neural networks,'' in \emph{NIPS}, 2012, pp. 1097--1105.

\bibitem{vggnet}
K.~Simonyan and A.~Zisserman, ``Very deep convolutional networks for
  large-scale image recognition,'' in \emph{ICLR}, 2015.

\bibitem{girshick2015fast}
R.~Girshick, ``Fast r-cnn,'' in \emph{ICCV}, 2015, pp. 1440--1448.

\bibitem{long2015fully}
J.~Long, E.~Shelhamer, and T.~Darrell, ``Fully convolutional networks for
  semantic segmentation,'' in \emph{CVPR}, 2015, pp. 3431--3440.

\bibitem{xie15hed}
S.~Xie and Z.~Tu, ``Holistically-nested edge detection,'' in \emph{ICCV}, 2015,
  pp. 1395--1403.

\bibitem{ren2015faster}
S.~Ren \emph{et~al.}, ``Faster r-cnn: Towards real-time object detection with
  region proposal networks,'' in \emph{NIPS}, 2015, pp. 91--99.

\bibitem{girshick2014rich}
R.~Girshick \emph{et~al.}, ``Rich feature hierarchies for accurate object
  detection and semantic segmentation,'' in \emph{CVPR}, 2014, pp. 580--587.

\bibitem{chen2016dcan}
H.~Chen \emph{et~al.}, ``Dcan: Deep contour-aware networks for accurate gland
  segmentation,'' in \emph{CVPR}, 2016, pp. 2487--2496.

\bibitem{jin2016object}
L.~Jin, Z.~Chen, and Z.~Tu, ``Object detection free instance segmentation with
  labeling transformations,'' \emph{arXiv preprint arXiv:1611.08991}, 2016.

\bibitem{Kainz2015Semantic}
P.~Kainz, M.~Pfeiffer, and M.~Urschler, ``Semantic segmentation of colon glands
  with deep convolutional neural networks and total variation segmentation,''
  \emph{arXiv preprint arXiv:1511.06919}, 2015.

\bibitem{sirinukunwattana2015stochastic}
K.~Sirinukunwattana, D.~R. Snead, and N.~M. Rajpoot, ``A stochastic polygons
  model for glandular structures in colon histology images,'' \emph{IEEE Trans.
  Medical Imaging}, vol.~34, no.~11, pp. 2366--2378, 2015.

\bibitem{Cheikh2016A}
B.~B. Cheikh, P.~Bertheau, and D.~Racoceanu, ``A structure-based approach for
  colon gland segmentation in digital pathology,'' in \emph{SPIE}, 2016, pp.
  97\,910J--97\,910J.

\bibitem{nguyen2012prostate}
K.~Nguyen, B.~Sabata, and A.~K. Jain, ``Prostate cancer grading: Gland
  segmentation and structural features,'' \emph{Pattern Recognition Letters},
  vol.~33, no.~7, pp. 951--961, 2012.

\bibitem{Li2016Gland}
W.~Li \emph{et~al.}, ``Gland segmentation in colon histology images using
  hand-crafted features and convolutional neural networks,'' in \emph{ISBI},
  2016, pp. 1405--1408.

\bibitem{ronneberger2015u}
O.~Ronneberger, P.~Fischer, and T.~Brox, ``U-net: Convolutional networks for
  biomedical image segmentation,'' in \emph{MICCAI}.\hskip 1em plus 0.5em minus
  0.4em\relax Springer, 2015, pp. 234--241.

\bibitem{xu2016gland}
Y.~Xu \emph{et~al.}, ``Gland instance segmentation by deep multichannel side
  supervision,'' in \emph{MICCAI}, 2016, pp. 496--504.

\bibitem{sermanet2013overfeat}
P.~Sermanet \emph{et~al.}, ``Overfeat: Integrated recognition, localization and
  detection using convolutional networks,'' in \emph{ICLR}, 2014.

\bibitem{ciresan2012deep}
D.~Ciresan \emph{et~al.}, ``Deep neural networks segment neuronal membranes in
  electron microscopy images,'' in \emph{NIPS}, 2012, pp. 2843--2851.

\bibitem{yu2015multi}
F.~Yu and V.~Koltun, ``Multi-scale context aggregation by dilated
  convolutions,'' \emph{arXiv preprint arXiv:1511.07122}, 2015.

\bibitem{lee2015deeply}
C.-Y. Lee \emph{et~al.}, ``Deeply-supervised nets,'' in \emph{AISTATS}, 2015,
  pp. 562--570.

\bibitem{jia2014caffe}
Y.~Jia \emph{et~al.}, ``Caffe: Convolutional architecture for fast feature
  embedding,'' in \emph{Proceedings of the 22nd ACM international conference on
  Multimedia}.\hskip 1em plus 0.5em minus 0.4em\relax ACM, 2014, pp. 675--678.

\bibitem{chen2016deeplab}
L.-C. Chen \emph{et~al.}, ``Deeplab: Semantic image segmentation with deep
  convolutional nets, atrous convolution, and fully connected crfs,''
  \emph{arXiv preprint arXiv:1606.00915}, 2016.

\end{thebibliography}

\end{document}